\documentclass[10pt,twocolumn,letterpaper]{article}

\usepackage{cvpr}              

\usepackage{graphicx}
\usepackage{amsmath}
\usepackage{amsthm}
\usepackage{amssymb}
\usepackage{booktabs}
\usepackage{comment}
\usepackage{multicol}
\usepackage{multirow}

\usepackage{color, colortbl}

\definecolor{LightCyan}{rgb}{0.88,1,1}

\usepackage[pagebackref,breaklinks,colorlinks]{hyperref}

\usepackage[capitalize]{cleveref}
\crefname{section}{Sec.}{Secs.}
\Crefname{section}{Section}{Sections}
\Crefname{table}{Table}{Tables}
\crefname{table}{Tab.}{Tabs.}

\def\cvprPaperID{8265} 
\def\confName{CVPR}
\def\confYear{2023}

\begin{document}

\title{Flow supervision for Deformable NeRF}

\author{
  Chaoyang Wang$^1$ \qquad
  Lachlan Ewen MacDonald$^{2}$ \qquad
  Laszlo A. Jeni$^1$ \qquad
  Simon Lucey$^{12}$ \vspace{2pt} \\
  $^1$Carnegie Mellon University \qquad
  $^2$University of Adelaide\\
  {\tt \small \{chaoyanw,laszlojeni\}@cs.cmu.edu \quad
  \{lachlan.macdonald, simon.lucey\}@adelaide.edu.au} \vspace{2pt}\\
  {\small \url{https://mightychaos.github.io/projects/fsdnerf}}
}

\maketitle
\def\c{\mathbf{c}} 
\def\C{\mathbf{C}} 
\def\P{\mathbf{p}} 
\def\traj{\mathcal{T}} 
\def\d{\mathbf{d}} 
\def\coeff{\boldsymbol{\varphi}}

\def\p{\mathbf{p}} 

\def\traj{\boldsymbol{\tau}}

\def\cI{\mathcal{I}}
\def\cT{\mathcal{T}}
\def\p{\mathbf{p}}
\def\t{\mathbf{t}}
\def\R{\mathbf{R}}
\def\D{\mathbf{D}}
\def\A{\mathbf{A}}
\def\B{\mathbf{B}}
\def\I{\mathbf{I}}
\def\supi{{(i)}}
\def\bd{\mathbf{d}}
\def\bz{\mathbf{z}}
\def\bw{\mathbf{w}}
\def\M{\mathbf{M}}
\def\S{\mathbf{S}}
\def\W{\mathbf{W}}
\def\w{\mathbf{w}}
\def\x{\mathbf{x}}
\def\t{\mathbf{t}}
\def\cW{\mathcal{W}}
\def\cL{\mathcal{L}}
\def\bvarphi{\boldsymbol{\varphi}}
\def\balpha{\boldsymbol{\alpha}}
\def\btheta{\boldsymbol{\theta}}
\def\blambda{\boldsymbol{\lambda}}
\def\bPsi{\mathbf{\Psi}}
\def\bpsi{\boldsymbol{\psi}}

\def\s{\mathbf{s}}
\def\Real{\mathbb{R}}
\def\so{\mathfrak{so}}
\def\SO{\mathbb{SO}}
\def\1{\mathbf{1}}
\def\xy{\text{xy}}
\def\Z{\mathbf{Z}}
\def\d{\mathbf{d}}
\def\C{\mathcal{C}}
\def\tW{\widetilde{\W}}
\def\tD{\widetilde{\D}}
\def\td{\widetilde{\d}}
\def\tbPsi{\widetilde{\bPsi}}
\def\tP{\widetilde{P}}

\def\idx{{(i)}}

\def\fwdw{w_{c\rightarrow}}
\def\bwdw{w_{c\leftarrow}}

\def\bwdwu{w_{c\leftarrow U}}

\newcommand{\SfM}{S\textit{f}M\xspace}
\newcommand{\SfC}{S\textit{f}C\xspace}
\def\SfMpp{S\textit{f}M++\xspace}

\newcommand{\centered}[1]{\begin{tabular}{l} #1 \end{tabular}}

\newcommand{\cy}{\textcolor{red}}
\begin{abstract}
In this paper we present a new method for deformable NeRF that can directly use optical flow as supervision. We overcome the major challenge with respect to the computationally inefficiency of enforcing the flow constraints to the backward deformation field, used by deformable NeRFs. Specifically, we show that inverting the backward deformation function is actually not needed for computing scene flows between frames. This insight dramatically simplifies the problem, as one is no longer constrained to deformation functions that can be analytically inverted. Instead, thanks to the weak assumptions required by our derivation based on the inverse function theorem, our approach can be extended to a broad class of commonly used backward deformation field. We present results on monocular novel view synthesis with rapid object motion, and demonstrate significant improvements over baselines without flow supervision.


\end{abstract}

\section{Introduction}
Reconstructing dynamic scenes from monocular videos is a significantly more challenging task compared to its static-scene counterparts, due to lack of epipolar constraints for finding correspondences and ambiguities between motion and structure. Recent advances in differentiable rendering have lead to various solutions using an analysis-by-synthesis strategy -- solving the non-rigid deformation and structure by minimizing the difference between synthesized images and input video frames. Among those, deformable neural radiance fields~\cite{Lombardi:2019,nerfies,dnerf,nr-nerf} has been a notable  technique to represent dynamic scenes and shows plausible space-time view synthesis results. However, the current implementations only warrant success on teleporting-like videos whose camera motions are significantly more rapid than object motions. Quality of their results significantly decrease on videos with more rapid object motions~\cite{gao2022monocular}. 

\begin{figure}[ht!]
    \centering
    \includegraphics[width=\linewidth]{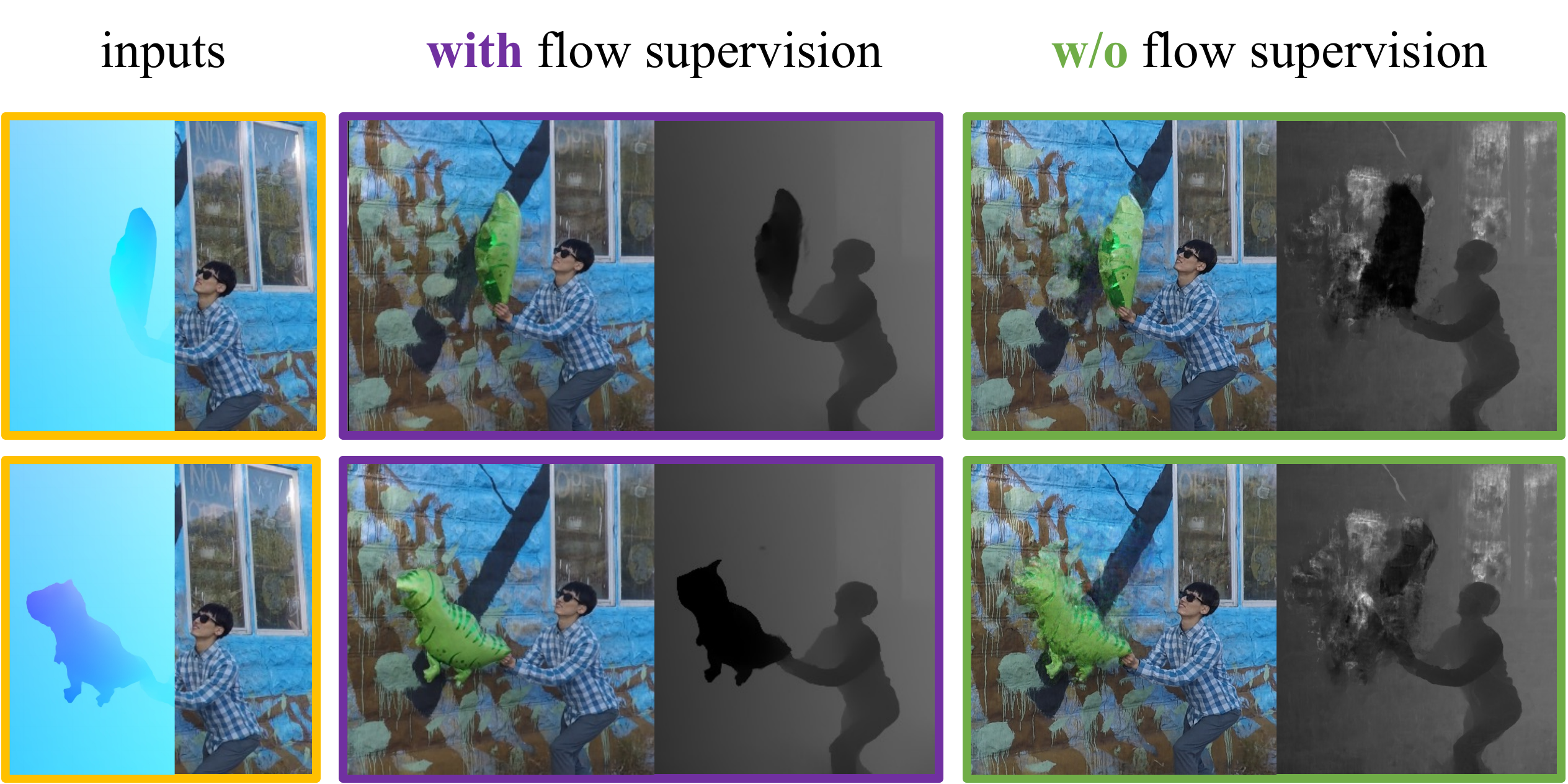}
    \vspace{-15pt}
    \caption{We propose a method to use optical flow supervision for deformable NeRF. It noticeably improves novel view synthesis for monocular videos with rapid object motions. In the figure, we visualize rendered novel view images and depth maps for the first and last frame of the input video.}
    \label{fig:teaser}
\end{figure}

In this work, we conjecture the deficiency of these deformable NeRF-based methods is mainly due to lack of temporal regularization. As they represent deformation as \emph{backward} warping from the sampled frames to some canonical space, the motions or scene flows between temporally adjacent frames is not directly modeled nor supervised. Another deficiency is that these methods minimize photometric error alone, which is insufficient for gradient descent to overcome poor local minima when the canonical and other frames has little spatial overlap. This deficiency is severe for non-object-centric videos with large translations. 

The community has explored optical flow as an additional cue to help supervise the temporal transitions of other motion representations, such as scene flow fields~\cite{flow-fields,gao2021dynamic,zhang2021consistent,du2021neural} and blend skinning fields~\cite{yang2021banmo}. However, enforcing flow constraints with respect to a generic \emph{backward} warping field as in Nerfies~\cite{nerfies} is non-trivial. Intuitively, to compute scene flows, it requires inverting the \emph{backward} warp by having a \emph{forward} warp which maps points from canonical space to other frames. Then scene flows can be computed either by taking time derivative of the forward warp or predicting the next position of a point through evaluating the forward warp. But this can be problematic since analytical inverse of a complicated non-bijective function (\eg neural networks) is impossible, and an approximate solution by having an auxiliary network to represent the forward warp will introduce computational overhead and is theoretically ill-posed. Counter to this intuition, we will show that inverting the backward warping function is actually not needed for computing scene flows between frames.  

The main \textbf{contribution} of this paper is:
we derive an analytical solution to compute velocities of objects directly from the \emph{backward} warping field of the deformable NeRF. The velocities are then used to compute scene flows through temporal integration, which allows us to supervise the deformable NeRF through optical flow. This leads to significant improvement on videos with more rapid motions, as shown in Fig.~\ref{fig:teaser}. 

The advantage of our approach is twofold:
(i) Our method applies to all kinds of backward warping function, thanks to the weak assumptions required by the inverse function theorem which our derivation is based on. Thus our method is more general compared to other works using invertible normalizing flows~\cite{lei2022cadex} or blend skinning~\cite{chen2021snarf,yang2021banmo}. (ii) Our method is also computationally more tractable compared to neural scene flow fields~\cite{flow-fields, du2021neural, li2021neural}, which would require integrating flows over long period of time to reach some canonical frame.

\section{Related works}
\noindent\textbf{Deformable NeRF.}
One common approach to model the dynamic scene is to represent it as the deformation of a static unknown template, and reconstruction is then done by fitting the model to the input 2D observations. Such approach has been implemented using techniques from recent advances in differentiable rendering. Most noticeably is deformable neural radiance field~\cite{Lombardi:2019,nerfies,dnerf,nr-nerf}, which models the deformation and the template radiance field using coordinate-based neural networks, and employs volumetric rendering to synthesis images under different viewing directions. 

More concretely, to synthesize the color of a pixel at time $t$, it first samples points along the line of sight. The sampled points $\p$'s are then fed into a \emph{backward} deformation field , \ie \begin{equation}
    \bwdw(\p; t) \longrightarrow \p_c, 
\end{equation}
which maps the input spacetime point $(\p, t)$ to its corresponding 3D position $\p_c\in\mathbb{R}^3$ in the canonical space. Colors $\mathbf{c}$ and densities $\sigma$ are then queried from the radiance field network, \ie
\begin{equation}
    f(\p_c, \mathbf{d}, \blambda) \longrightarrow \mathbf{c}, \sigma,
\end{equation}
where $\mathbf{d}\in S^2$ is the viewing direction and $\blambda$ is an additional frame-wise code to
allow the template to vary per frame so as to cope with topological and appearance changes~\cite{nerfies,nerf-w}. $\blambda$ can also be extended to vary spatially as ambient embeddings to enable greater flexibility to topological changes~\cite{hypernerf}. With the computed colors and densities of points along the viewing ray, RGB intensities of a pixel are computed using the volumetric rendering equations proposed in NeRF~\cite{Mildenhall20eccv_nerf}. Finally, the optimization objective is to minimize the difference between rendered pixels and the input observations.

\noindent\textbf{Backward deformation fields.}
Different ways exist for representing the backward deformation field. The most straightforward design is to use a neural network to output displacement between the input point and its canonical position~\cite{derf,nr-nerf,Lombardi:2019}. Park~\etal shows that having the neural network outputting SE(3) transformations leads to improvement in reconstructing rotational motions~\cite{nerfies}. 

For articulated objects such as animals and humans, blend skinning is widely used in literature~\cite{kavan2014part,loper2015smpl,zuffi20173d,zhi2020texmesh}. However, it is mainly designed for \emph{forward} deformation, thus adjustment is needed to adapt it for \emph{backward} warping field. Yang \etal~\cite{yang2021banmo,yang2021viser} uses mixture of gaussians to model the blending weights. This enables them to approximate the inverse of a forward blend skinning by simply inverting the SE(3) transformation of each deformation nodes. On the other hand, instead of explicitly define a deformation function, Chen~\etal~\cite{chen2021snarf} solves for canonical correspondences of any deformed point using iterative root finding.

For homeomorphic deformation where the mappings between any frames are bijective and continuous, recent works explored the use of invertible normalizing flows~\cite{lei2022cadex,cai2022neural} where the forward and backward deformation are computed with the same network parameters. The downside is the network architecture is restrictive, and in practice requires more compute due to having more coupling layers to enumerate different axis partitions.

\noindent\textbf{Other dynamic NeRF representations.}
Instead of having a static template NeRF, other works~\cite{flow-fields,wang2021neural,gao2021dynamic} choose to use a time-modulated NeRF to directly represent warped radiance fields. To enforce temporal consistency, they optimize neural scene flow fields to regularize pairwise motion between adjacent frames. This is only suitable for enforcing short-term consistency, but intractable for long-term consistency due to the expensive computational cost for performing scene flow integration. To improve computational efficiency, Wang~\etal~\cite{wang2021neural,Wang_ntp_2022_CVPR} propose neural trajectory field which directly outputs trajectories for all space-time locations. This allows enforcing long-term consistency without the need for scene flow integration.

\noindent\textbf{Optical flow supervision.}
Using optical flows to assist view interpolation~\cite{Wang_2022_CVPR,huang2020rife,bao2019depth} and 3D reconstruction~\cite{consistentdetph,kopf2021robust,teed2018deepv2d} has been a common practise. Several recent dynamic NeRF works~\cite{du2021neural,flow-fields,wang2021neural,gao2021dynamic} also use optical flow supervision, but none of them is based on backward deformation fields. Yang ~\etal~\cite{yang2021banmo} apply optical flow to supervise a blend skinning deformation field for object-centric reconstruction. Their result focuses on articulated objects and requires amodel segmentation mask as input. Under the context of view synthesis for dynamic scenes, we are unaware of any deformable NeRF-based approach using flow supervision. Thus our result provides a useful tool for future related research.

\begin{figure}
    \centering
\includegraphics[width=.95\linewidth]{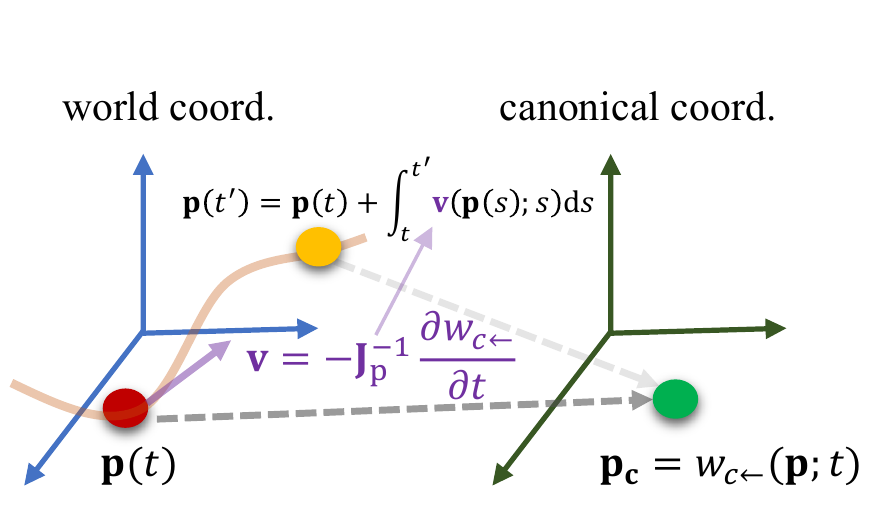}
    \vspace{-5pt}
    \caption{Given a point at $\p(t)$ and a \emph{backward} warping field $\bwdw$, we want to compute the 3D scene flow by predicting the next position of the point at time $t'$. We achieve this by deriving the velocity field $v$ as a differentiable function of $\bwdw$, and then perform time integration.}
    \label{fig:illu}
\end{figure}

\section{Supervise deformable NeRF by flow}
\noindent\textbf{Problem setup.} We concern the problem of fitting the deformable NeRF given a monocular video input. We precomputed the camera intrinsics and extrinsics using off-the-shelf structure-from-motion methods \eg Colmap~\cite{colmap}. Optical flows between neighboring frames $o_{t\rightarrow t\pm \nabla t}$ are also computed using RAFT~\cite{raft}. Then we want to find optimal parameters for the backward deformation field $\bwdw$ and radiance field $f$ such that the synthesised images $c_t'$ and optical flow maps $o_{t\pm t'}'$ match the input video frames $c_t$ and precomputed optical flow maps $o_{t\pm t'}$,
\begin{equation}
\resizebox{0.9\hsize}{!}{$
   \min \sum_{t} \left[ \underbrace{\|c_t' - c_t\|_2^2}_{\text{\small{image loss}}} + \beta \sum_{t'\in\{t \pm \Delta t\}} \underbrace{\|M_{t\rightarrow t'} \odot ( o_{t\rightarrow t'}' - o_{t\rightarrow t'})\|_1}_{\text{\small{optical flow loss}}} \right]$}
  \label{eq:train_obj}
\end{equation}
To prevent errors in the precomputed optical flow maps from misleading the reconstruction, we use binary masks $M_{t\rightarrow t'}$ to turn off losses for pixels which fail the forward-backward flow consistency test. Moreover, we follow the trick proposed by Li \etal~\cite{flow-fields} to gradually decay the weighting $\beta$ during optimization, so that the reconstruction is able to correct mistakes of the input optical flows.

The key question is then how to synthesize optical flows $o_{t\rightarrow t'}'$ from $\bwdw$ and $f$?  

\begin{figure}
    \centering
    \includegraphics[width=.95\linewidth]{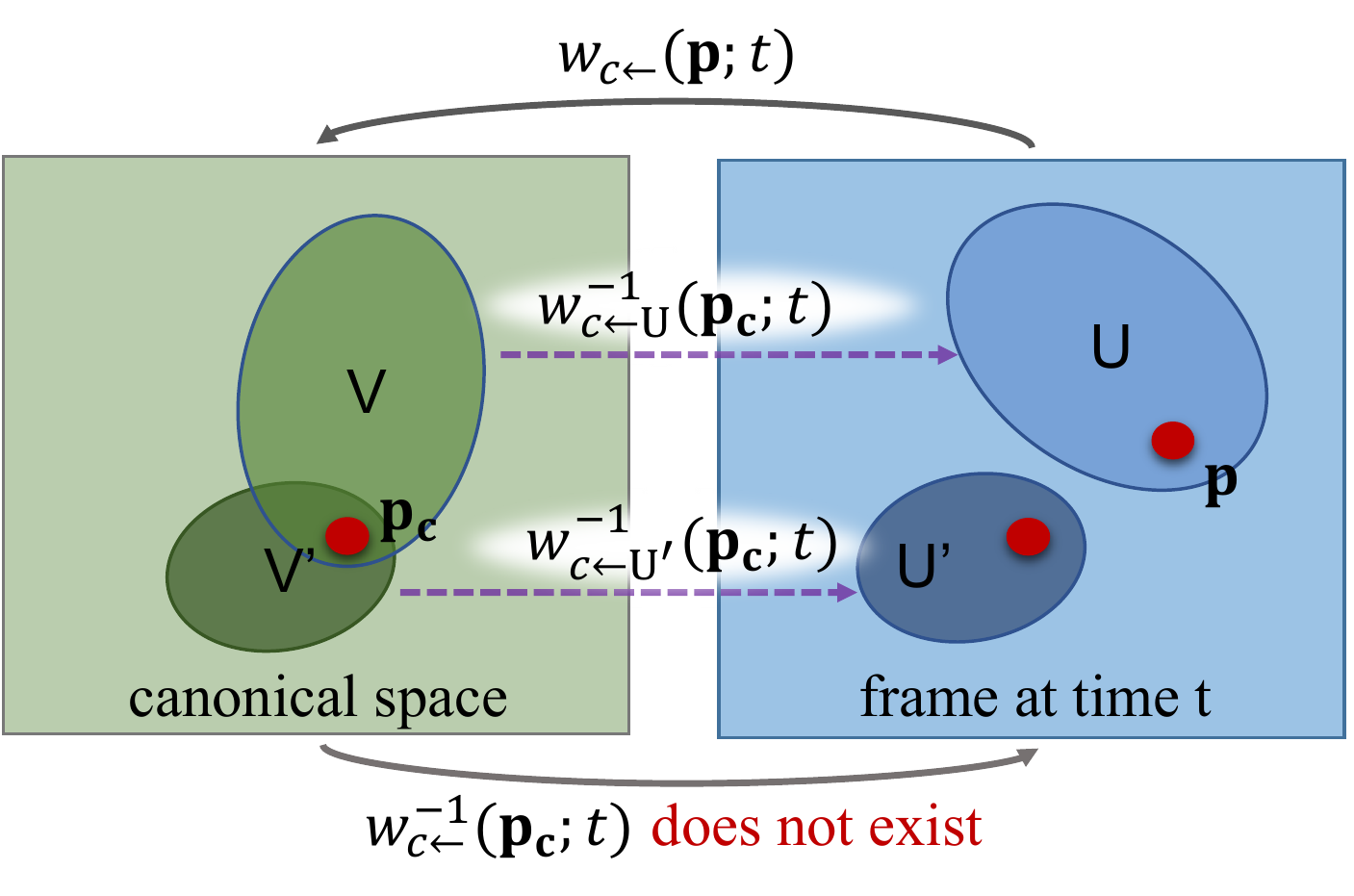}
    \vspace{-5pt}
    \caption{Illustration of the invertibility for \emph{backward} deformation field $\bwdw$. Modeled by coordinate-based MLP, $\bwdw(\p;t)$ is not invertible on the whole input domain. But local homeomorphism exists. we can have a bijective mapping $\bwdwu(\p;t)$ from an open set $U$ to another open set $V$ in the canonical space. Although $\bwdwu^{-1}(\p_c;t)$ exists locally from $V$ to $U$, but it is not possible to analytically derive it in closed form. Fortunately, inverting $\bwdwu$ is not needed for predicting scene flow due to equation (\ref{eq:velocity}). }
    \label{fig:inv_func_th}
\end{figure}

\subsection{Velocity fields from $\bwdw (p;t)$}
Velocity fields $v(\p;t):\mathbb{R}^3\times \mathbb{R} \rightarrow \mathbb{R}^3$ describe the velocity of an object if placed at position $\p$ in the world coordinate at time $t$. 
Velocity fields is straightforward to compute if the \emph{forward} deformation field $\fwdw(\p_c;t):\mathbb{R}^3\times \mathbb{R} \rightarrow \mathbb{R}^3$ exists, \ie:
\begin{equation}
    v(\p;t) = \frac{\partial \fwdw}{\partial t}\bigg|_{(\bwdw(\p;t);t)}.
\end{equation}
However, this intuitive solution is flawed if $\fwdw$ is not strictly an inverse function of $\bwdw$. Unfortunately having an invertible $\bwdw$ is theoretically unwarranted since most deformation representations such as neural networks and  blend skinnings are not bijective.

Since seeking a \emph{global} inverse function of $\bwdw$ on the whole domain is impractical, we propose to consider local regions of $\bwdw$'s domain where \emph{local} inverse function may exist. And based on the inverse function theorem, we find that we can analytically compute $v(\p;t)$ for any positions $\p$ without actually inverting $\bwdw$, as long as $\bwdw$ is bijective in an open set includes $\p$. This leads to the main theoretical result of the paper.

\noindent\textbf{Proposition.} If the warp Jacobian matrix $\mathbf{J}_\p (\p,t) = [\frac{\partial \bwdw(\p;t)}{\partial\p}]$ is non-singular at some position $\p$ at time $t$, and there exists an open set including $\p$ where $\bwdw$ is continuously differentiable, then the velocity at $(\p, t)$ is computed as:
\begin{equation}
    v(\p;t) = - \mathbf{J}_\p^{-1}(\p,t) \frac{\partial \bwdw (\p;t)}{\partial t}.
\label{eq:velocity}
\end{equation}
\begin{proof} First, we upgrade $\bwdw$ to have the same input and output dimension, \ie $\phi(\p, t) = (\bwdw(\p;t), t)$. Then given the assumptions made by the proposition, and according to the inverse function theorem~\cite{munkres2018analysis}, $\phi$ is a local diffeomorphism. In other words, there is some open set $U$ containing $(\p,t)$ and an open set $V$ containing $\phi(\p,t)$ such that $\phi_U := \phi: U\rightarrow V$ has a continuous and differentiable inverse $\phi_U^{-1} : V \rightarrow U$; in particular, for $(\p_c,t) = \phi(\p,t)$, we have $(J\phi_U^{-1})(\p_c,t) = [(J\phi_U)(\p,t)]^{-1}$. Re-expressing $\phi_U = (\bwdw, t)$ and denoting the inverse of $\bwdw$ inside the open set $U$ as $\bwdwu^{-1}$ (see Fig.~\ref{fig:inv_func_th}) yields :
\begin{equation}
\resizebox{0.9\hsize}{!}{$
    \begin{bmatrix} J_{\p} \bwdwu^{-1} &
    \frac{\partial \bwdwu^{-1}}{\partial t}\\
    \mathbf{0}_3^\top & 1\end{bmatrix} \bigg|_{(\p_c,t)}= 
    \begin{bmatrix}
    J_\mathbf{p} \bwdw &
    \frac{\partial \bwdw}{\partial t}\\
    \mathbf{0}_3^\top & 1
    \end{bmatrix}^{-1}\bigg|_{(\p,t)} $}
\label{eq:jac_ift}
\end{equation}
By moving the matrix inverse inside the block matrix on the righthand side of (\ref{eq:jac_ift}) through Schur complement, we have $\frac{\partial \bwdwu^{-1}(\p_c,t) }{\partial t} = -[(J_\p\bwdw)(\p,t)]^{-1}\frac{\partial \bwdw (\p,t)}{\partial t}$. Finally by noticing that $v(\p;t) = \frac{\partial \bwdwu^{-1}(\p_c,t) }{\partial t}$, we have the proof. 
\end{proof}
We note that the sufficient condition of the proposition is weak and satisfied for most domain of deformation fields. For rare cases where $\det (\mathbf{J}_\p) < \epsilon$ for some space-time positions $(\p,t)$, we choose to exclude it from evaluating the loss so as to avoid numerical instability. Moreover, we implement equation (\ref{eq:velocity}) as a differentiable operator so that gradients can be back propagated during optimization.

\subsection{Scene flow from time integration of velocity}
With the velocity fields $v(\p;t)$ computed by equation (\ref{eq:velocity}), we estimate 3D scene flows \ie the displacement between $\p(t)$ and $\p(t+\Delta t)$ by time integration (see Fig.~\ref{fig:illu}),
\begin{equation}
    \p(t+\nabla t) - \p(t) = \int_t^{t+\Delta t} v(\p(s);s)ds.
    \label{eq:t_int}
\end{equation}
We implement the time integration though differentiable numerical ODE solvers. Because in our problem we are dealing with scene flows between small time intervals $\nabla t$, which usually is the duration of 1 or 2 frames, we find unrolling the fourth order Runge-Kutta method~\cite{hamming2012numerical} for two iterations gives stable results and achieves good trade-off between accuracy and computational cost.
To prevent overfitting to a fixed step size, we randomly perturbed the step size by adding a Gaussian noise.

\noindent\textbf{Rendering optical flow.} For every viewing ray at time $t$, we first calculate the next position $\p(t+\Delta t)$ for each sampled points along the ray by equation (\ref{eq:t_int}). Then the expected next position $\bar{\p}(t+\Delta t)$ for the visible points is estimated by weighted averaging $\p(t+\Delta t)$'s using NeRF's volumetric rendering equation. Finally, the optical flow is estimated by taking the difference between the 2D projection of $\bar{\p}(t+\Delta t)$ and the pixel location of the viewing ray.

\begin{figure}
    \centering
    \includegraphics[width=\linewidth]{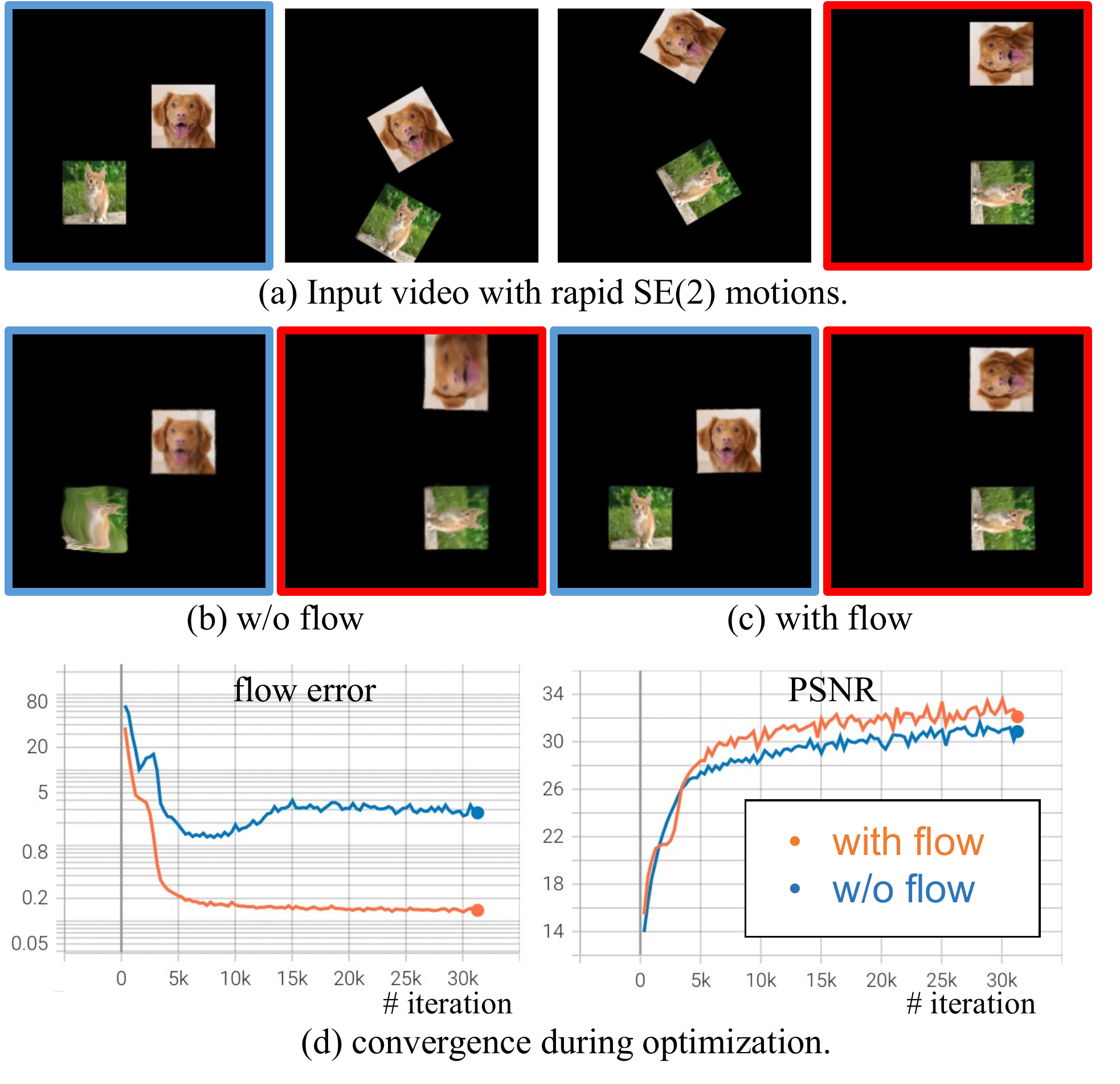}
    \vspace{-20pt}
    \caption{A 2D toy experiment showing flow supervision is crucial for reconstructing rapid object motions. (a) 4 frames evenly sampled from the synthetic video. Two image patches are moving with large linear and angular velocity and has no overlap between their starting and ending positions. (b) fitting a SE(2) deformation field fails to recover correct motions for the first and last frame. (c) adding flow supervision correctly reconstruct the video frames.(d) we monitor the flow error and PSNR of reconstructed image during  optimization. The flows are estimated using equation (\ref{eq:velocity},\ref{eq:t_int}). Our method (with flow) converges significantly faster than w/o flow, and is able to fit the input optical flows with low error.}
    \label{fig:toy_exp}
\end{figure}

\subsection{Removing Gauge freedom}
One issue for the deformable NeRF is the recovered background tends to be not static. Deformation of the apparent static scene regions severely affects viewing experience. We find that a main cause for the jittering background is due to the Gauge ambiguity in deformable NeRF, \ie the optimization objective in (\ref{eq:train_obj}) does not specify which canonical coordinate the deformation field is defined at. In theory, any deformation of a canonical space can also be a valid one. This causes confusion for the deformable NeRF during optimization, which tends to trap it in the wrong factorization of motion and shapes.

Therefore, we propose to remove the Gauge freedom by specifying that the canonical coordinate is attached to a key frame $t_0$ from the input sequence. This is enforced through adding a loss to penalize the magnitude of deformation at time $t_0$, \ie
\begin{equation}
    \mathcal{L}_\text{Gauge} = \sum_\p \|\bwdw(\p;t_0) - \p\|_2.
\end{equation}
 Through our experiments, we choose the middle frame of the input sequence as the canonical frame. This is based on the assumption that the average deformation to other frames is likely to be minimal in the middle frame. Nevertheless, more sophisticated algorithms of choosing canonical frames may further improve the robustness of the method. 
 

We note that prior works would introduce extra regularization to enforce static background.  Compared to training the density field to align with sparse points estimated by structure from motion~\cite{nerfies,spacetime-nerf}, our approach is self-contained without external modules. Compared to prior works having separate NeRFs for static and dynamic regions~\cite{nerf-w,flow-fields,gao2021dynamic,wu2022d}, our implementation is computationally more friendly. Though all the above methods could be included as supplementary.  

\begin{figure*}[t]
    \centering
    \includegraphics[width=\linewidth]{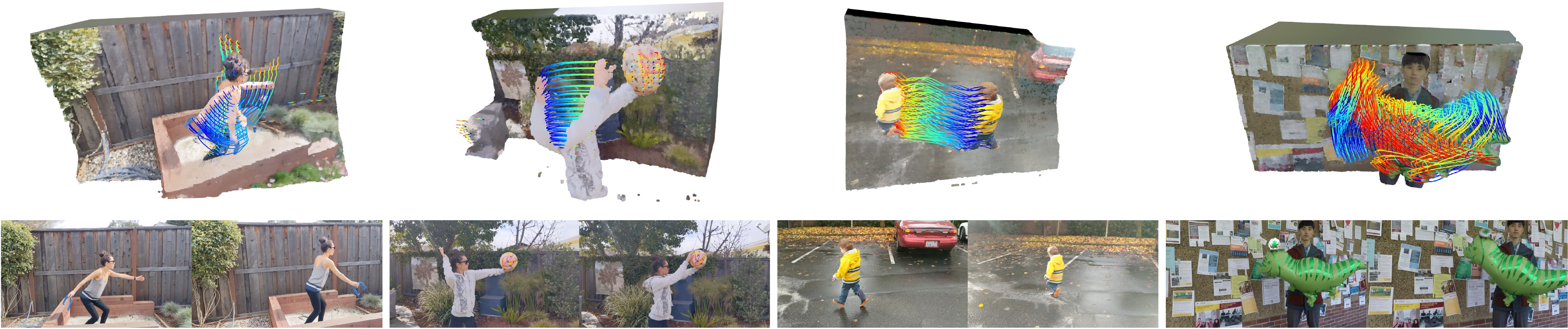}
    \caption{3D visualizing of trajectories computed by integrating velocity fields $v(\p;t)$ from equation (\ref{eq:velocity}). Trajectories are colored to represent temporal order, and overlayed with colored point clouds extracted from the optimized deformable NeRF. Bottom row shows two frames from each of the input videos.}
    \label{fig:traj_teaser}
\end{figure*}

\section{Experiment}

\subsection{Implementation details}
\noindent\textbf{Deformation field.} We follow Nerfies~\cite{nerfies} by using a 6 layer 128-width MLP which outputs SE(3) transformations. The deformation $\text{MLP}_g$ takes input $\p \in \mathbb{R}^3$ and an 8-dim deformation code. Instead of treating the codes as independent variables, we use another 2 layer 32-width $\text{MLP}_\alpha$ to map time $t$ to the deformation code. This helps improve convergence under smaller batch size. We also choose to use softplus as the activation function, since it produces smoother outputs than ReLU. With these, mathematically $\bwdw(\p;t) = \text{MLP}_g(\p, \text{MLP}_\alpha(t)) \circ \p$, where $\circ$ denotes the action of SE(3) transform.

\noindent\textbf{Radiance field.} We use the original architecture from Nefies with minor modification. Instead of having independent 8-dim appearance codes to optimize, we use a 2 layer 32-width MLP to produce the codes conditioned on time $t$.

\noindent\textbf{Optimization hyperparameters.} We set the initial weighting $\beta$ for the optical flow loss as 0.04 and anneal it to 0.0001. We set a fixed weighting for $\mathcal{L}_\text{Gauge}$ as 1. The initial learning rate of Adam optimizer is 0.001, and decayed 1/10 every 50 epochs. We train a total of 120 epochs or roughly 150k iterations. Each batch samples 4096 rays from 8 frames, and samples 256 points per ray. The optimization takes 4 GTX-3090 GPUs approximately 13hrs.

\subsection{Effectiveness of flow supervision}
In experiments, we evaluate the effectiveness of flow supervision using the velocity fields derived in equation ~(\ref{eq:velocity}). We aim to answer two key qustions: 
\begin{itemize}
    \item \emph{Does our method help improve convergence for rapid motions?}
    \item \emph{Does the flow supervision help disambiguate dynamic motion and structure in monocular deformable NeRF?}
\end{itemize}
\subsubsection{Flow supervision for fitting rapid motions}
To clearly address the first question, we conduct a 2D toy experiment. As shown in Fig.~\ref{fig:toy_exp}, we created a 25-frame video, consists of two rapidly moving image patches. Unlike the synthetic experiment in Nerfies~\cite{nerfies}, the image patches have large translational motion in addition to rotations, and the motions are different for each patches. It turns out that fitting an SE(2) deformation field using only image intensity loss is slow to converge, and results in distorted images. Applying optical flow loss by our method significantly speeds up convergence, and gives correct image reconstruction. This result indicates that our flow calculation algorithm \ie equation~(\ref{eq:velocity},\ref{eq:t_int}) is effective and also flow supervision is necessary for handling rapid motions. 

\subsubsection{Monocular dynamic view synthesis}
\noindent\textbf{Dataset.} State-of-the-art deformable NeRF, \eg Nerfies and HyperNeRF~\cite{nerfies,hypernerf} have shown satisfactory view synthesis result on the data they captured. However as discussed by Gao \etal~\cite{gao2022monocular}, Nerfies and HyperNeRF's data have high effective multi-view factors (EMFs). In other words, the objects are either quasi static or the camera motions are significantly larger than object motions. In this work, we evaluate on datesets with less EMFs. 

We first report results on the NVIDIA dynamic view synthesis dataset (NDVS)~\cite{globalcoherent} which has significantly lower EMFs. We follow the preprocessing steps of NSFF~\cite{flow-fields} which extracted 24 frames per sequence from the raw multi-view videos in NDVS. Neighboring frames are extracted from different cameras to simulate a monocular moving camera. For fair comparison with NSFF, we also downsize the images to have 288 pixels in height. For evaluation, we compare synthesized images to all images captured by 12 cameras, and report metrics such as PSNR, SSIM~\cite{wang2004image} and LPIPS~\cite{zhang2018perceptual}.

We next compare view synthesis results on sequences collected by the authors of Nerfies~\cite{nerfies}. To ensure the inputs appear as if they were casually recorded in real life, we use video frames only from the left camera from the stereo rig, as opposed to teleporting between the left and right cameras in the original paper of Nerfies.

We also test our method on one DAVIS sequence~\cite{Perazzi2016} and two casual videos captured by Wang \etal~\cite{wang2021neural}.

\noindent\textbf{Trajectories by velocity integration.} To visually inspect the quality of our optimized deformation field and the derived velocity fields, we sparsely sample points on the surface of the reconstructed scene, and perform time integration with the velocity fields to create trajectories. As visualized in Fig.~\ref{fig:traj_teaser}, the recovered trajectories are smooth and closely follow the object motions. 

\begin{figure}
    \centering
    \includegraphics[width=\linewidth]{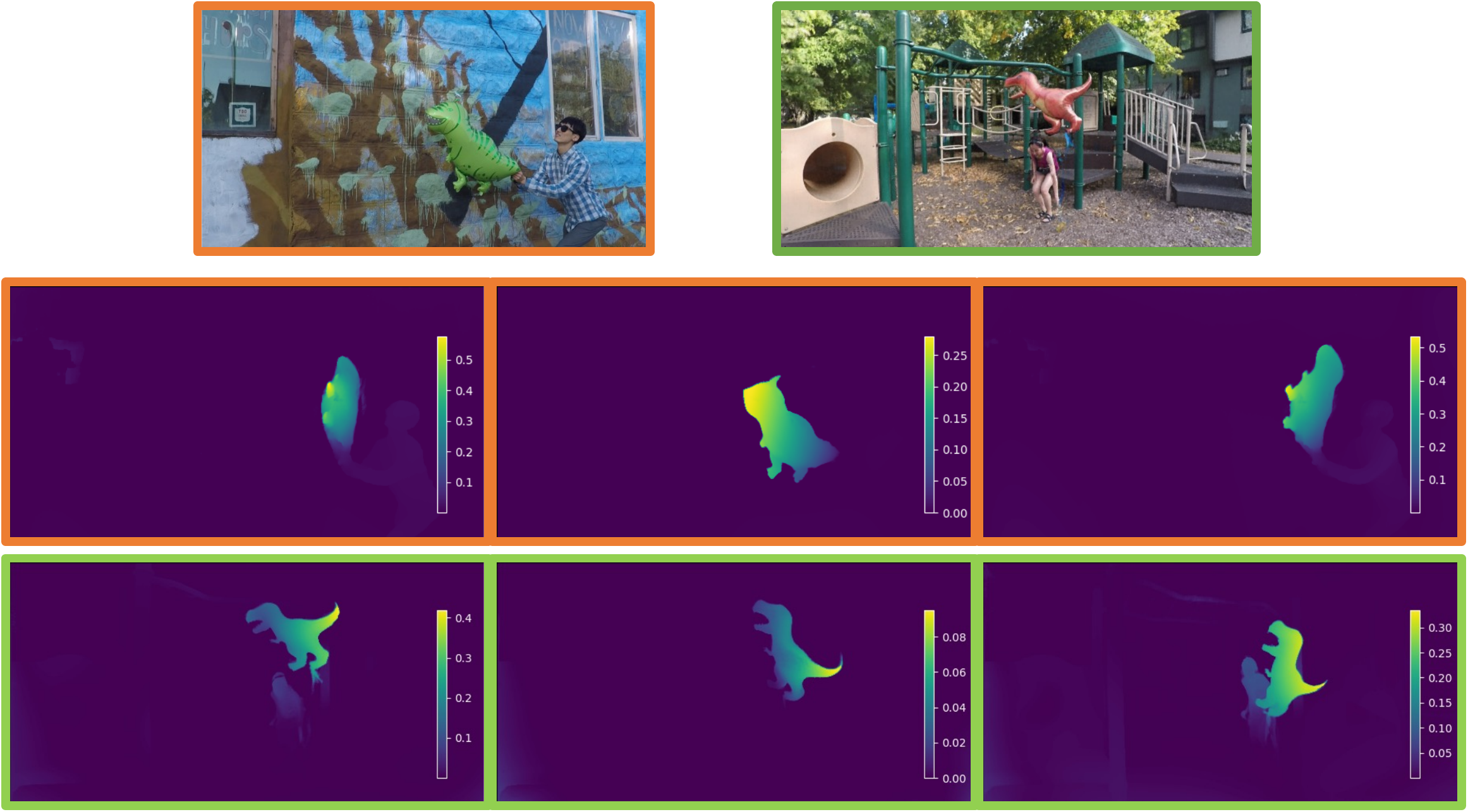}
    \vspace{-15pt}
    \caption{Visualization the distance $\|\p-\p_c\|_2$ by volumetric rendering. Since we removed the Gauge freedom by picking the canonical frame to be one of the input frame, large distance only happens on the moving objects. Our result shows clean separation between the moving forground objects and the static background.}
    \label{fig:distance_teaser}
\end{figure}

\noindent\textbf{Foreground background separation.} Since we removed Gauge freedom by picking one video frame as the canonical frame, the distance of a point $\p$ to its canonical correspondence $\p_c$ now directly indicates whether the point is static or moving. In Fig.~\ref{fig:distance_teaser}, we visualized the distance $\|\p-\p_c\|_2$ for each frame in a video. To visualize 3D volumes of distances in 2D, we project the distances along a ray by the volumetric rendering equation in NeRF. Thus brightness of the color corresponds to the distance of the visible area. We only observe large distances for the moving balloons and some small distances for the human subjects. The distances on the static background region are correctly rendered as close to 0. This indicates our success on the proposed removal of Gauge freedom.

\noindent\textbf{Baselines.}
We formed a close comparison with the state-of-the-art deformable NeRFs \eg Nerfies~\cite{nerfies} on the NDVS dataset. Due to the official code from the authors do not work out of the box for the NDVS dataset, we adapt it by changing its Euclidean coordinates to the NDC coordinates, so as to automatically deal with the increased scene depth in some of the NDVS sequences. Given all the aforementioned implementation and design choices, our own method is essentially applying the proposed flow supervision to the adapted Nerfies implementation. Thus comparison to Nerfies also serves as an ablation showing the effectiveness of the proposed flow supervision.

We also compared with another deformable NeRF approach, \ie NR-NeRF~\cite{nr-nerf}, whose deformation network outputs translation rather than SE(3) transformation. Finally, as a reference, we compared to NSFF~\cite{flow-fields}, which optimizes a time-modulated NeRF and scene flow fields, and is supervised not only by optical flows but also by the depth maps from the state-of-the-art monocular depth estimation network~\cite{midas}. Thus NSFF serves as a strong reference to check the status of  other methods without depth supervision.
We summarize the compared methods in Table.~\ref{tab:setup}. 
\begin{figure*}
    \centering
    \includegraphics[width=\linewidth]{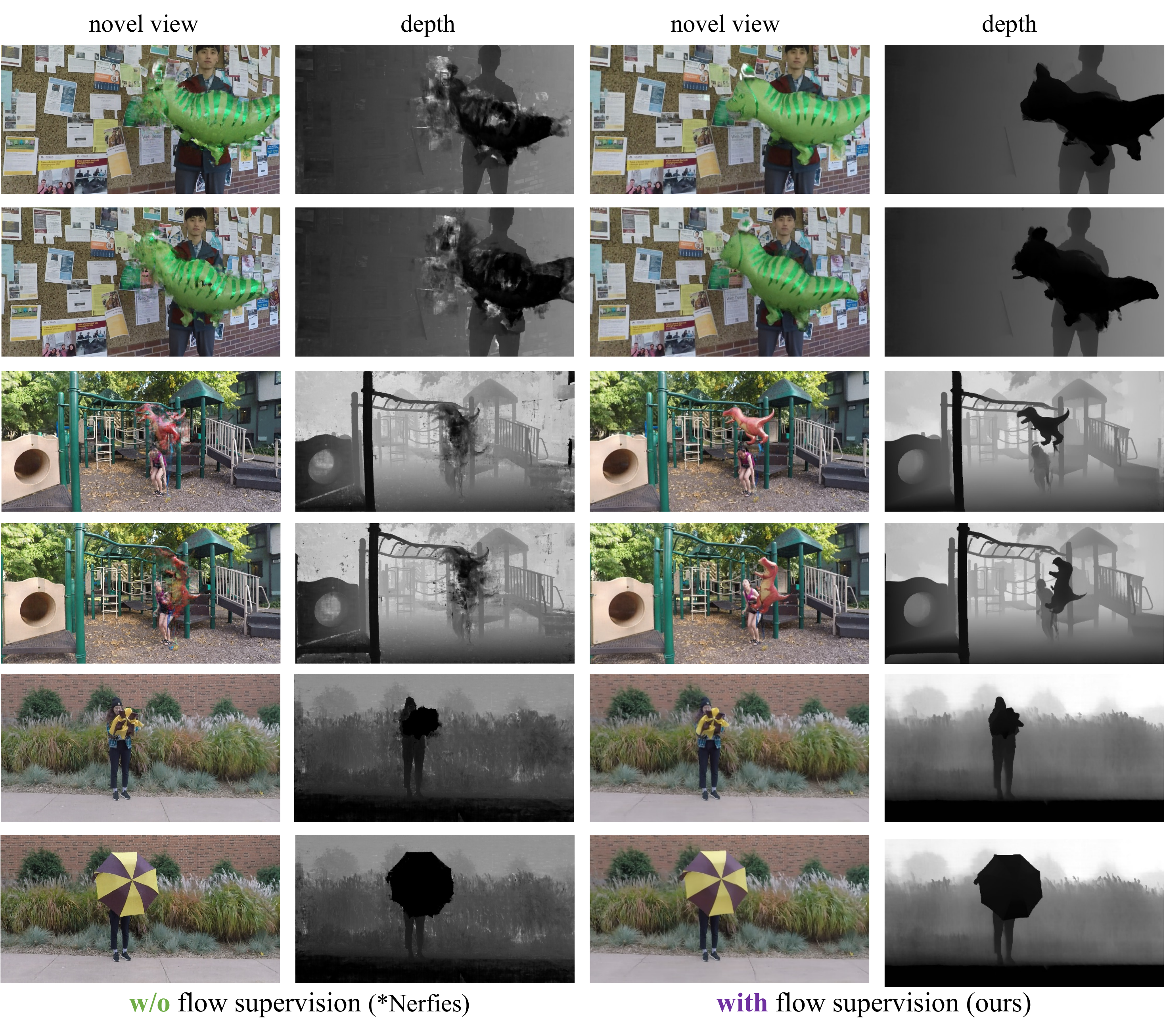}
    \vspace{-25pt}\caption{Results on NVIDIA dyanamic view synthesis dataset (NDVS). With flow supervision, our method produces smooth depth map and sharp novel view images. Without flow supervision leads to severe artifacts and noisy depth maps. We note that *Nerfies is our own adaptation of the official code for NDVS dataset.}\label{fig:main_result}
    \vspace{-5pt}
\end{figure*}

\begin{figure}
    \centering
     \includegraphics[width=\linewidth]{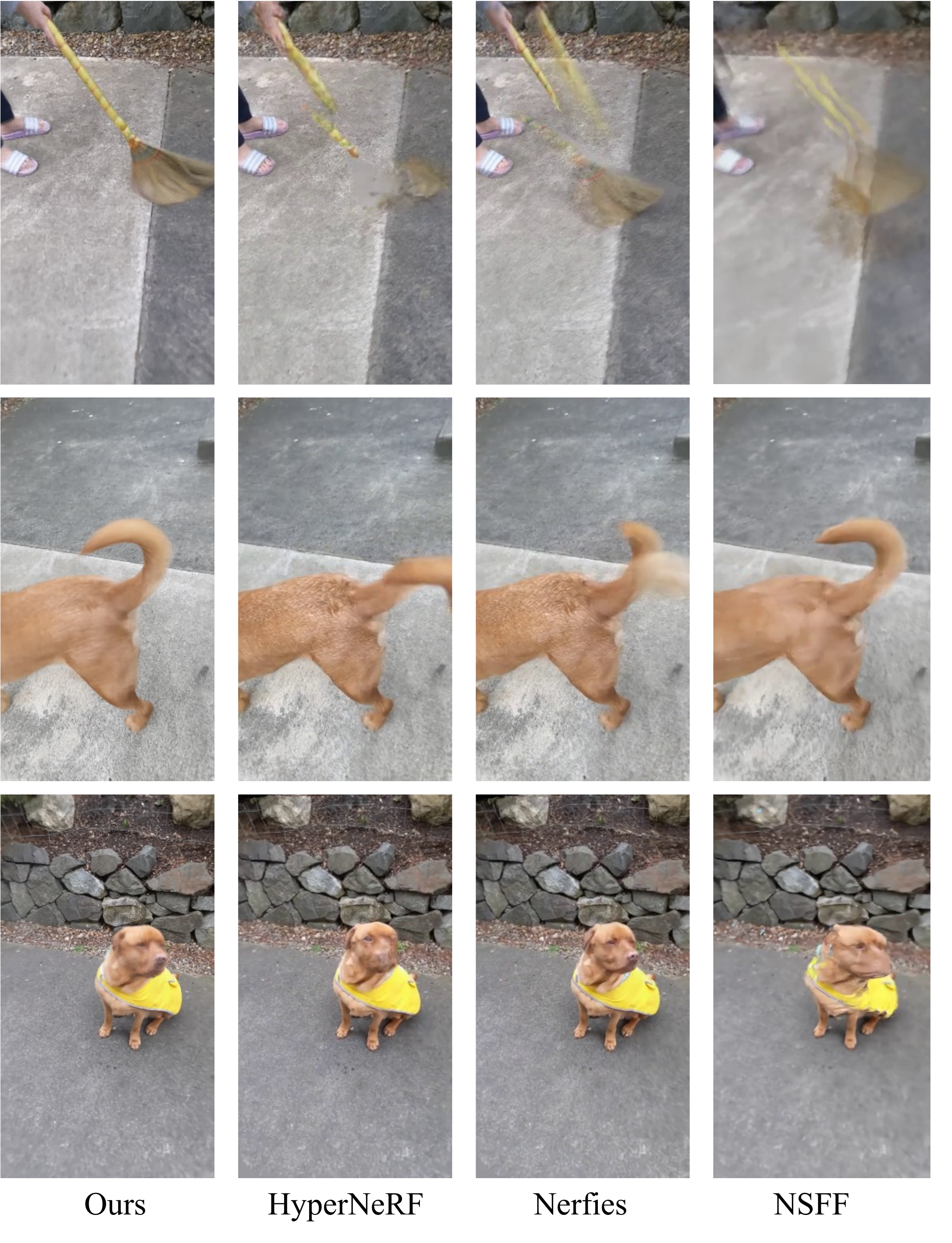}
    \vspace{-20pt}
    \caption{View synthesis results on sequences from Nerfies~\cite{nerfies}. We only use frames from the left camera for training, instead of teleportation between left and right cameras as in the original paper of Nefies. We find the compared methods make noticeable mistakes in the ``broom'' sequence (1st row) and some frames in the ``tail'' sequence (2nd row). In the "toby-sit" sequence (3rd row), HyperNeRF and NSFF produce blurry or distorted dog faces in some frames. In contrast, our method consistently yields a more plausible view synthesis result. This indicates that adding flow supervision by our method is also helpful for quasi-static videos.}
    \label{fig:nerfies}
    \vspace{-25pt}
\end{figure}

\begin{table}[h!]
    \centering
    \resizebox{\hsize}{!}{
    \begin{tabular}{c|cc|ccc}
    \toprule
     & \multicolumn{2}{c}{representation} & \multicolumn{3}{c}{supervision}\\
    method & motion & NeRF & image & flow & depth \\
    \hline
    NSFF~\cite{flow-fields} & scene flow & dynamic  & \checkmark & \checkmark & \checkmark \\
    \hline
    NR-NeRF~\cite{nr-nerf} & translation & static & \checkmark & & \\
    Nerfies~\cite{nerfies} & SE(3) & static & \checkmark & &\\
    HyperNeRF~\cite{hypernerf} & SE(3) + ambient & semi-static & \checkmark & &\\
    ours & SE(3) & static & \checkmark & \checkmark &\\
    \bottomrule
    \end{tabular}}
    \vspace{-2pt}
    \caption{Summary of the compared deformable NeRF methods and neural scene flow field (NSFF).}
    \label{tab:setup}
\end{table}

\begin{table*}[h]
    \centering
\resizebox{\hsize}{!}{
    \begin{tabular}{cc|ccc|ccc|ccc|ccc}
    \toprule
        method &   & \multicolumn{3}{c}{Playground} &
        \multicolumn{3}{c}{Balloon1} &
        \multicolumn{3}{c}{Balloon2} &
        \multicolumn{3}{c}{Umbrella}\\
         & & PSNR$\uparrow$ & SSIM$\uparrow$ & LPIPS$\downarrow$ & PSNR$\uparrow$ & SSIM$\uparrow$ & LPIPS$\downarrow$ & PSNR$\uparrow$& SSIM$\uparrow$ & LPIPS$\downarrow$ & PSNR$\uparrow$ & SSIM$\uparrow$ & LPIPS$\downarrow$ \\
    \hline
    NSFF~\cite{flow-fields} & full & 24.69 & 0.889 & 0.065 & 24.36 & 0.891 & 0.061 & 30.59 & 0.953 & 0.030 & 24.40 & 0.847 & 0.088\\
    & dyn. & 19.02 & 0.715 & 0.123 & 18.49 & 0.619 & 0.174 & 24.46 & 0.843 & 0.065 & 16.82 & 0.546 & 0.156\\
    \hline
    \rowcolor{LightCyan}
    \multirow{2}{*}{NR-NeRF~\cite{nr-nerf}} &  full & 14.16 & 0.337 & 0.363 & 15.98 & 0.444 &  0.277 & 20.49 & 0.731 & 0.348 & 20.20  &  0.526 & 0.315 \\
    & dyn. & 11.78 &  0.221 & 0.466 & 16.94 & 0.548 &  0.398 & 12.65 & 0.353 &  0.575 & 16.20  & 0.435 & 0.321\\
    \rowcolor{LightCyan}
    w/o flow & full & 22.18 &  0.802 &  0.133 & 23.36 &  0.852 &  0.102 & 24.91 & 0.864 & 0.089 & 24.29 & 0.803 & 0.169 \\
    (*Nerfies) & dyn. & 16.33 & 0.535 &  0.244 & 18.66 & 0.613 & 0.215 & 20.50 & 0.717 & 0.141 & 17.57 & 0.581 & 0.202\\
    \rowcolor{LightCyan}
    w flow & full & 22.39 & 0.812 & \textbf{0.109} & 24.36 & 0.865 & 0.107 & 25.82 & 0.899 & \textbf{0.081} & 24.25 &  0.813 & \textbf{0.123}\\
    (ours)                & dyn. & 16.70 & 0.597 & \textbf{0.168} & 19.53 & 0.654 & \textbf{0.175} & 20.13 & 0.719 & \textbf{0.113} & 18.00 & 0.597 & \textbf{0.148}\\
    \bottomrule
    \end{tabular} }
    \caption{Comparing deformable NeRFs and NSFF on the NVIDIA dynamic view synthesis (NDVS) dataset. Metrics are reported for the full image as well as only for the masked regions containing dynamic motions. Our method shows significant improvement over deformable NeRF methods without flow supervision. Comparison with NSFF is mixed, mainly due to the scale ambiguity without depth supervision. See Fig.~\ref{fig:scale_issue} for further discussion.}
    \vspace{-10pt}\label{tab:ndvs_result}
\end{table*}

\noindent\textbf{With vs. w/o flow supervision.} In Fig.~\ref{fig:main_result} we form a side-by-side comparison with Nerfies, which does not use optical flow supervision. We notice that Nerfies constantly make structural mistakes as indicated by its rendered noisy depth maps. As a result, it has noticeable artifacts concentrated around the dynamic objects. In contrast, with the help from flow supervision, our method renders smoother depth maps and visually more pleasing view synthesis images. These improvements are reflected quantitatively in Tab.~\ref{tab:ndvs_result}, where we show consistent improvement across all metrics. Our method also produces more plausible results for quasi-static scenes as shown in Fig.~\ref{fig:nerfies}.

\begin{figure}
    \centering
    \includegraphics[width=\linewidth]{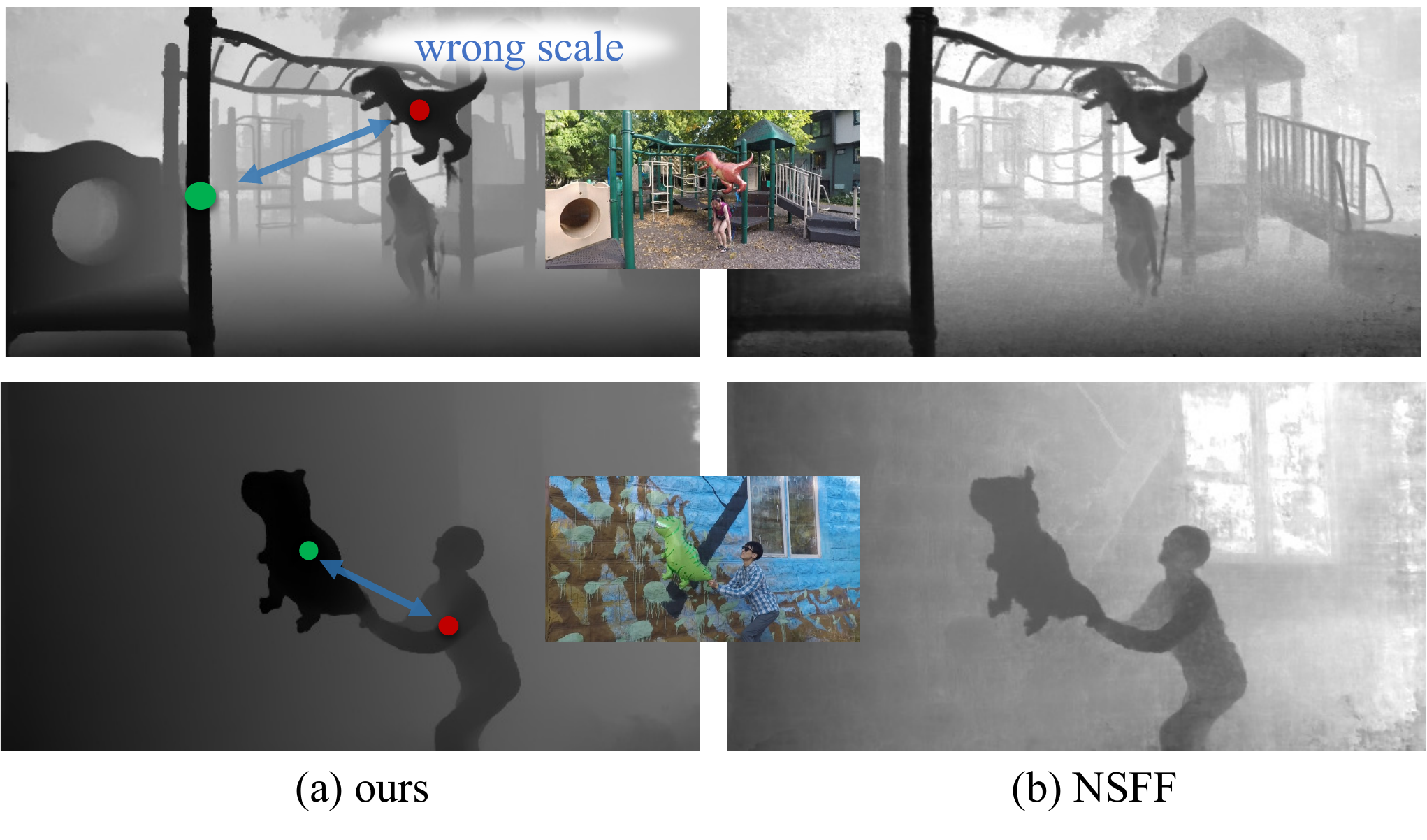}
    \vspace{-20pt}
    \caption{Our method suffers from scale ambiguity due to not using depth supervision. This is highlighted on the top left by comparing the points on the balloon and the pole, where the balloon should be behind the pole rather than having similar depth. Similarly, on the bottom left, the arm of the person should be close to the balloon, not behind it. Although we produce much smoother depth maps compared to NSFF, we make more error in the scale of depth, resulting in lower metrics in Table~\ref{tab:ndvs_result}. }
    \label{fig:scale_issue}
\end{figure}

\noindent\textbf{Compare with NSFF.} In table~\ref{tab:ndvs_result} we show competitive results on Balloon1 and Umbrella compared to NSFF which is supervised using depth. However we fall behind on the Playground and Balloon2 sequence. Closer diagnoses show that this is due to wrong depth scales which our method assigned to the fast moving balloons. As shown in Fig.~\ref{fig:scale_issue}, our method actually produces smoother depth maps compared to NSFF, thanks to our stronger temporal constraint due to having a single static template NeRF. However as highlighted by the blue arrows, the depth value of the moving objects are too small in comparison to the reference points on the background. We hypothesize this is due to the small motion bias of the deformation field which tends to explain 2D motions with smaller 3D motions. This causes the rendered foreground objects have significant offsets compared to the groundtruth, and consequently receives large penalties in terms of the image similarity metrics used in Table~\ref{tab:ndvs_result}, even though our method produces equivalent if not sharper view synthesis result compared to NSFF. This scale ambiguity issue is inherent from the single camera problem setup and should not blame the methods supervised without depth. To recover plausible relative scales of different moving parts of a dynamic scene, mid or higher level reasoning (\eg learning-based depth estimation~\cite{midas,wsvd}, 2D supervision from image generative models~\cite{poole2022dreamfusion}) is required.

\begin{figure}
    \centering
    \includegraphics[width=\linewidth]{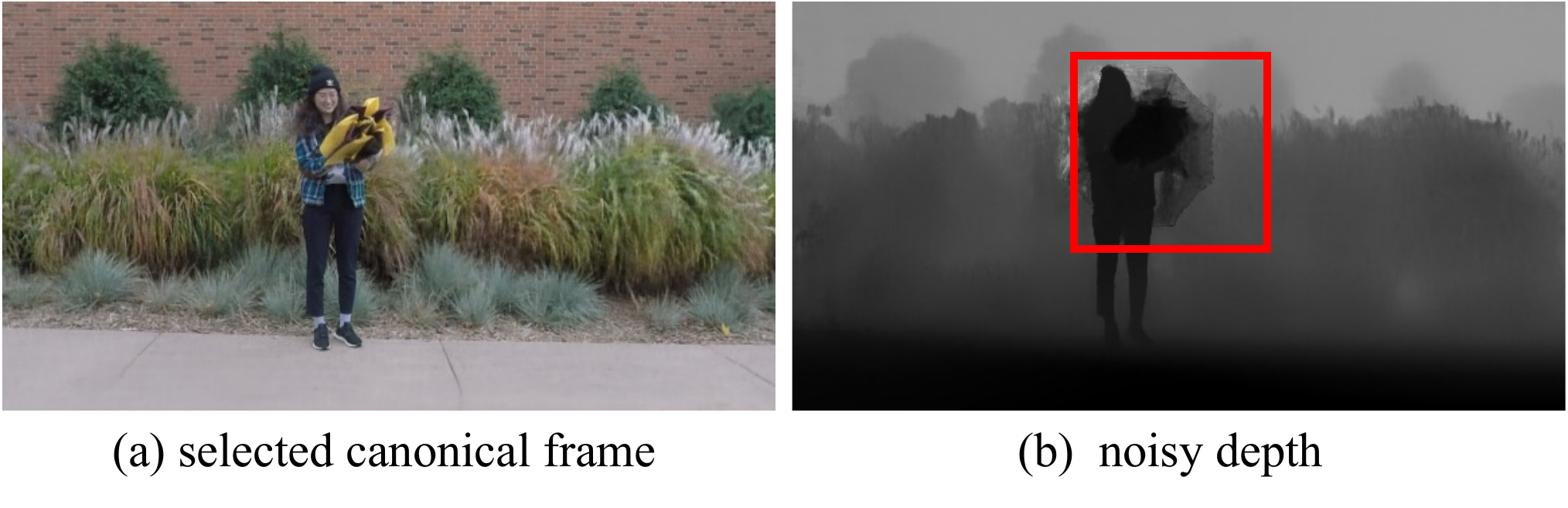}
    \vspace{-20pt}
    \caption{For highly deformable objects such as umbrella, the choice of the canonical frame is sensitive. In this example, we choose the first frame instead of mid frame as the canonical frame, which has a very different topogy compared to other frames when the umbrella is open. This leads to degenerated results as highlighted by the \textcolor{red}{red} box. }
    \label{fig:can_issue}
    \vspace{-10pt}
\end{figure}
\section{Discussion}
We presented a method to apply flow supervision for deformable NeRF. We demonstrated that our method significantly improves view synthesis quality of deformable NeRF on videos with lower effective multi-view factors. However, due to the ambiguities of the monocular 3D reconstruction problem, our method has following \textbf{limitations:} (i) our method is not able to recover correct relative scale of moving objects (see Fig.~\ref{fig:scale_issue}); (ii) our method can be sensitive to the selection of canonical frame if there is large object deformations (see Fig.~\ref{fig:can_issue}); (iii) it requires sufficient motion parallax and does not work for fixed or small camera motion; (iv) Long optimization time is required, but could be sped up using more efficient implementation \cite{mueller2022instant}. 

\noindent\textbf{Acknowledgement.} This work was partially supported by Apple and by NSF award No. IIS-1925281. We thank Tim Clifford and Ian R Fasel from Apple for the helpful discussions, Orazio Gallo for collecting data and Hang Gao for helping with the baselines.

\appendix
\section{Implementation details}
\subsection{SE(3) transformation in normalized device coordinate}
For unbounded frontal scenes in the DVS dataset~\cite{globalcoherent}, normalized device coordinate (NDC)~\cite{Mildenhall20eccv_nerf} is required to squeeze unbounded 3D space into a bounded one with respect to the depth direction. To perform SE(3) transformation of points with the NDC coordinates, naive implementation would be first convert the NDC coordinates to Euclidean coordinates, apply the SE(3) transformation, and then convert it back to the NDC coordinates. However doing so may run into numerical instability issue for points in long distance.  Alternatively we derived a formula to directly apply SE(3) transformation in the NDC coordinates. For any NDC coordinates $(x,y,z)$, its corresponding NDC coordinates $(x',y',z')$ after applying the SE(3) transformation with rotation $\mathbf{R}\in SO(3)$ and translation $\mathbf{t}\in \mathbb{R}^3$ is calculated as follow:
\begin{align}
    \begin{pmatrix} a \\ b\\ c \end{pmatrix} & = \mathbf{R} \begin{pmatrix} \frac{2f_x x}{W} \\ \frac{2f_y y}{H}\\ 1 \end{pmatrix} + \frac{1-z}{2n} \mathbf{t} \\
    \begin{pmatrix} x' \\ y'\\ z' \end{pmatrix} & =  \begin{pmatrix} 
        \frac{2f_x a}{W c} \\ 
        \frac{2f_y b}{H c}\\ 
        \frac{1+c-z}{c} \end{pmatrix},
\end{align}
where $f_x, f_y, H, W, n$ are the focal lengths, image sizes and near plane distance as in the definition of NDC coordinates. The above formulation avoids the hassle of dealing with points at infinity in the Euclidean coordinates. 

\subsection{Efficient implementation of equation~(5)}
To be able to backpropagate through equation~(5) and make the computation tractable for evaluating on 100k+ points per iteration, we made the following implementation design choices. 

First, we choose to implement a 2nd-order differentiable quaternion-based operator using CUDA. This is 100x faster than implementation with native pytorch operators as in PyTorch3D~\cite{ravi2020pytorch3d}, and 5x faster than matrix multiplication based implementation. We note that LieTorch library~\cite{teed2021tangent} also provides fast implementation of SE(3) transformation and computes gradients in the tangent space. However it does not support 2nd-order derivatives which is required for optimization with respect to the velocities estimated by equation~(5).

For calculating the $3\times 3$ matrix inverse of the Jacobian matrices, me choose to implement an analytical solution of the matrix inverse with CUDA, which is 100x faster than pytorch's generic matrix inverse routine when dealing with 100k $3\times 3$ small matrices. We use the following equation to calculate the derivative,
\begin{equation}
    \frac{\partial \mathbf{H}^{-1}}{\partial h_{ij}} = - \mathbf{H}^{-1} \boldsymbol{\delta}_{ij}\mathbf{H}^{-1},
\end{equation}
where $\boldsymbol{\delta}_{ij}$ is a  binary matrix whose $(i,j)$th entry is the element with value of 1.
In our CUDA implementation, we parallalize the compute for each element of $\frac{\partial \mathbf{H}^{-1}}{\partial h_{ij}}$ so as to be significantly faster than the generic matrix inverse operators from pytorch.

\section{Comparison to directly inverting the backward deformation field.} 
\begin{figure}
 \includegraphics[width=\linewidth]{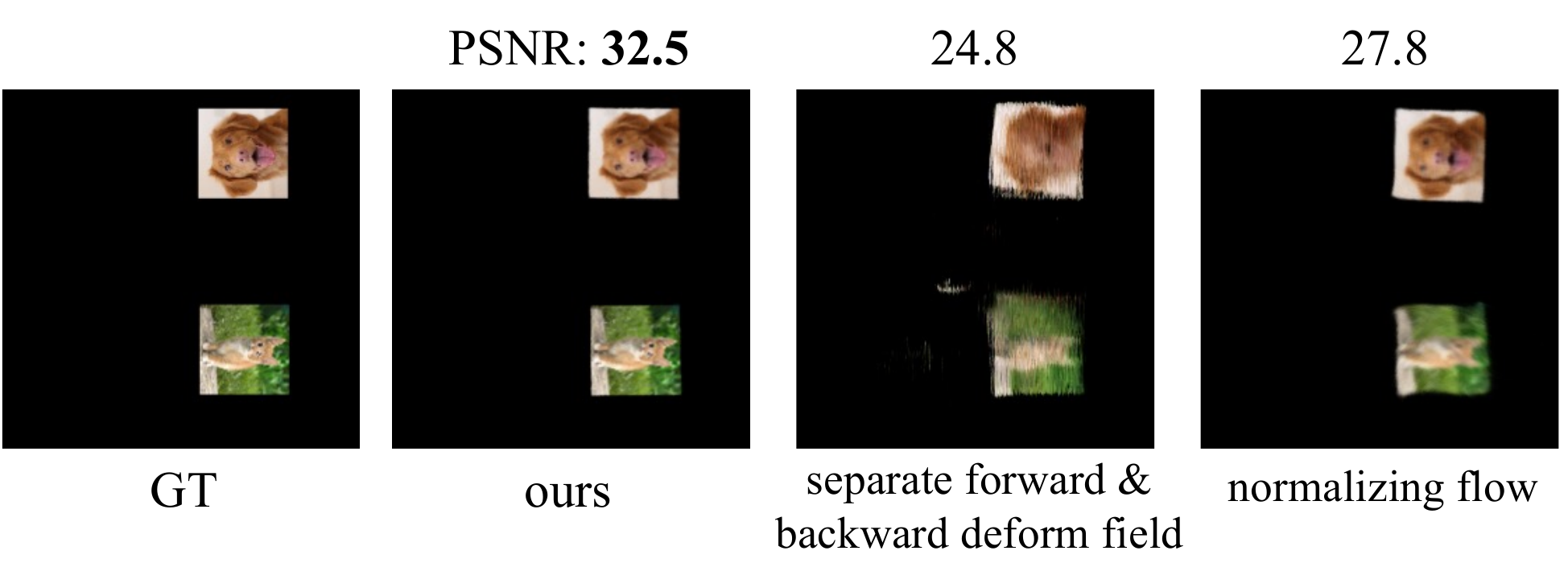}
\caption{Compare flow supervision using different deformation representation. Our method (2nd column) outperforms baselines using separate forward \& backward deformation field (3rd column) and bijective normalizing flow (4th column).} 
\label{fig:ablation_invertible}
\end{figure}

In Fig.~\ref{fig:ablation_invertible}, we compare our method against two baselines which directly invert the backward deformation field. The experiment setup is to fit a video with se(2) motions as described in Figure 4 of the main paper. The optimization objective is to minimize the combination of photometric loss and optical flow loss.

The first baseline (3rd column) has a separate MLP which represents the \emph{forward} deformation field $\fwdw$. It has the same architecture as the \emph{backward} deformation field $\bwdw$. Then the optical flow between frame $t$ and $t+\Delta t$ at pixel $\p$ is estimated as $\mathbf{o}_{t\rightarrow t+\Delta t} = \fwdw(\bwdw(\p,t),t+\Delta t) - \p$. As shown in the 3rd column of the following figure, this baseline is not able to reconstruct the input video with high fidelity (PSNR=24.8). This is due to there being no explicit gaurantee that the forward and backward deformation fields are cyclic consistent. 

The 2nd baseline applies normalizing flow as in Lei \& Daniilidis~\cite{lei2022cadex} The deformation field is modeled by Real-NVP, which is a bijective mapping with analytic inverse. We find that due to the network architecture restrictions, Real-NVP is not able to perfectly fit the motions (see distortion of image patches in the 4th column), even when it has significantly more layers than our method (12 vs 4).

In comparison, our method achieves the highest PSNR which indicates its effectiveness against directly inverting the backward deformation field. Finally, we note that it may be possible to make the baselines stronger by carefully tuning loss functions or network architectures, however the main strength of our method is its generality without the need for introducing additional modules or tuning.  

\section{Other optical flow rendering/loss alternatives}
In our preliminary investigation, we also experimented different ways of rendering the optical flow from 3D scene flows and evaluating the flow loss. All of the following alternatives are inferior to the one we described in the main paper, and we have recorded them here for reference.

\begin{itemize}
    \item The first alternative is instead of evaluating scene flows for all sampled points along the ray and then weighted averaging them, we only evaluate scene flows for a single point on the estimated visible surface. More specifically, we first synthesize the depth as in NeRFs~\cite{nerfies,Mildenhall20eccv_nerf,flow-fields} by weighted averaging depth of sampled points along the ray. This gives an estimation of a position $\mathbf{p}(t)$ on the visible surface. Then the next position $\mathbf{p}(t+\Delta t)$ is estimated by equation (7). 
    \item We still evaluate scene flows for all the points along the ray. But instead of aggregate them together to form a single optical flow, we project all scene flows to 2D flows, and directly evaluate optical flow loss by comparing them to the input optical flow. To account for occlusions, we weighted the loss by the weights from the volumetric rendering equation. More specifically, given the projected 2D flows $\mathbf{o}_i\in\mathbb{R}^2$ for each point $i$ along the ray, we evaluate the optical flow loss as follow:
    \begin{equation}
        \mathcal{L_\text{o.f.}} = \sum_i w_i\|\mathbf{o}_i - \mathbf{o}_{\text{input}}\|_1,
    \end{equation}
    where $w_i$ denotes the weights from the volumetric rendering and $\mathbf{o}_{\text{input}}\in\mathbb{R}^2$ is the input optical flow.
\end{itemize}

In our current experiment, we found both above approaches are inferior to the one we used in the main paper. Sometimes they resulted in training instability or floating surfaces. A deeper understanding of why these two approaches do not work may motivate more efficient approaches of enforcing the flow loss.

{\small
\bibliographystyle{ieee_fullname}
\bibliography{egbib}
}

\end{document}